\documentclass[11pt,a4paper]{article}
\usepackage[hyperref]{naaclhlt2019}
\usepackage{times}
\usepackage{latexsym}
\usepackage{amsfonts}
\usepackage{amsmath}
\usepackage{graphicx}
\usepackage{xspace}
\usepackage{multirow}

\usepackage{url}

\aclfinalcopy % Uncomment this line for the final submission
\def\aclpaperid{497} %  Enter the acl Paper ID here

\def\HRERE{{\sc Hrere}\xspace}

\title{Connecting Language and Knowledge with Heterogeneous Representations for Neural Relation Extraction}

\author{Peng Xu \\
  Department of Computing Science \\
  University of Alberta\\
  Edmonton, Canada\\
  {\tt pxu4@ualberta.ca} \\\And
  Denilson Barbosa \\
  Department of Computing Science \\
  University of Alberta\\
  Edmonton, Canada\\
  {\tt denilson@ualberta.ca} \\}

\date{}

\begin{document}
\maketitle
\begin{abstract}
Knowledge Bases (KBs) require constant updating to reflect changes to the world they represent.
For general purpose KBs, this is often done through Relation Extraction (RE), the task of predicting KB relations expressed in text mentioning entities known to the KB. 
One way to improve RE is to use KB Embeddings (KBE) for link prediction.
However, despite clear connections between RE and KBE, little has been done toward properly unifying these models systematically.
We help close the gap with a framework that unifies the learning of RE and KBE models leading to significant improvements over the state-of-the-art in RE.
The code is available at \url{https://github.com/billy-inn/HRERE}.
\end{abstract}

\section{Introduction}
%!TEX root = ../naaclhlt2019.tex

Knowledge Bases (KBs) contain structured information about the world and are used in support of many natural language processing applications such as semantic search and question answering.
Building KBs is a never-ending challenge because, as the world changes, new knowledge needs to be harvested while old knowledge needs to be revised.
This motivates the work on the Relation Extraction (RE) task, whose goal is to assign a KB relation to a \emph{phrase} connecting a pair of entities, which in turn can be used for updating the KB.
The state-of-the-art in RE builds on neural models using distant (a.k.a. weak) supervision~\cite{mintz2009distant} on large-scale corpora for training.

\begin{figure*}[ht]
\begin{center}
 \includegraphics[height=1.7in]{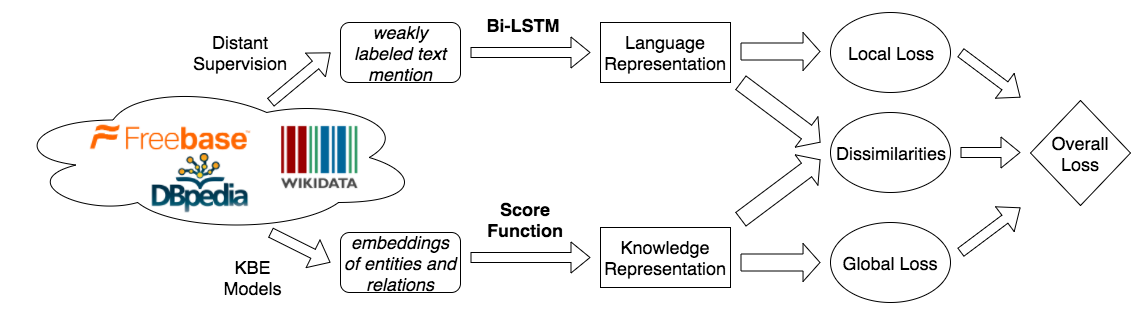}
\end{center}
\caption{Workflow of the proposed framework.} \label{workflow}
\end{figure*}

A task related to RE is that of Knowledge Base Embedding (KBE), which is concerned with representing KB entities and relations in a vector space for predicting missing links in the graph.
Aiming to leverage the similarities between these tasks, \citet{weston2013connecting} were the first to show that \emph{combining} predictions from RE and KBE models was beneficial for RE.
However, the way in which they combine RE and KBE predictions is rather naive (namely, by adding those scores).
To the best of our knowledge, there have been no systematic attempts to further unify RE and KBE, particularly during model \emph{training}. 

We seek to close this gap with \HRERE (Heterogeneous REpresentations for neural Relation Extraction), a novel neural RE framework that learns language and knowledge representations \emph{jointly}.
Figure~\ref{workflow} gives an overview.
\HRERE's backbone is a bi-directional long short term memory (LSTM) network with multiple levels of attention to learn representations of text expressing relations. 
The knowledge representation machinery, borrowed from {\bf ComplEx}~\cite{trouillon2016complex}, nudges the language model to agree with facts in the KB.
Joint learning is guided by three loss functions: one for the language representation, another for the knowledge representation, and a third one to ensure these representations do not diverge.
In effect, this contributes to \HRERE's generalization power by preventing over-fitting by either model.

We build on state-of-the-art methods for learning the separate RE and KBE representations and on learning tools that allow us to scale to a moderately large training corpus. (We use a subset of Freebase with 3M entities as our KB.)
We validate our approach on an established benchmark against state-of-the-art methods for RE, observing not only that our base model significantly outperforms previous methods, but also the fact that jointly learning the heterogeneous representations consistently brings in improvements.
% We also show the contribution of the different ingredients of our approach, and illustrate in particular the benefits of using the cross-entropy loss function for joint model learning.
To the best of our knowledge, ours is the first principled framework to combine and jointly learn heterogeneous representations from both language and knowledge for the RE task.

\paragraph*{Contributions.}
This paper describes and evaluates a novel neural framework for jointly learning representations for RE and KBE tasks that uses a cross-entropy loss function to ensure both representations are learned together, resulting in significant improvements over the current state-of-the-art for the RE task.

\section{Related Work}
%!TEX root = ../naaclhlt2019.tex

% \subsection{Neural Relation Extraction}
%\paragraph*{Neural Relation Extraction.}

Recent neural models have been shown superior to approaches using hand-crafted features for the RE task.
Among the pioneers, \newcite{zeng2015distant} proposed a piecewise convolutional network with multi-instance learning to handle weakly labeled text mentions.
Recurrent neural networks (RNN) are another popular architecture \cite{wu2017adversarial}. 
%Various neural models are proposed to improve the performance by different techniques, including attention mechanism \cite{jiang2016relation,lin2016neural}, adversarial learning \cite{wu2017adversarial}, deep residual learning \cite{huang2017deep}, incorporating relation path \cite{zeng2016incorporating}, to name a few.
% \subsection{Knowledge Base Embedding}
% \paragraph*{Knowledge Base Embedding.}
Similar fast progress has been seen for the KBE task for representing entities and relations in KBs with vectors or matrices.
\citet{bordes2013translating} introduced the influential translation-based embeddings (TransE), while
\citet{yang2014embedding} leveraged latent matrix factorization in their DistMult method.
We build on ComplEx \cite{trouillon2016complex}, which extends DistMult into the complex space and has been shown significantly better on several benchmarks.
%There are more models proposed in this area \cite{wang2014transH, lin2015learning, ji2015knowledge, xiao2015one, nickel2016holographic, dettmers2017convolutional} to help learn better representations of KB.

% \subsection{Connecting Text Information and Knowledge Base}
% \paragraph*{Connecting Text and Knowledge Base Representations.}

% {\bf To facilitate relation extraction}:
\citet{weston2013connecting} were the first to connect RE and KBE models for the RE task. Their simple idea was to  train the two models \emph{independently} and only combine them at inference time.
While they showed that combining the two models is better than using the RE model alone, newer and better models since then have obviated the net gains of such a simple strategy \cite{xu2018investigations}. 
We propose a much tighter integration of RE and KBE models: we not only use them for prediction, but also \emph{train} them together, thus mutually reinforcing one another.
% combination schemes common in the literature such as linear combinations, geometric mean or harmonic mean~\cite{peng2018invest}.

Recently, many methods have been proposed to use information from KBs to facilitate relation extraction.
\newcite{sorokin2017context} considered other relations in the sentential context while predicting the target relation.
\newcite{vashishth2018reside} utilized additional side information from KBs for improved RE.
However, these methods didn't leverage KBE method to unify RE and KBE in a principled way.
\newcite{han2018neural} used a mutual attention between KBs and text to perform better on both RE and KBE, but their method was still based on TransE \cite{bordes2013translating} which can not fully exploit the advantage of the information from KBs.

% {\bf To facilitate knowledge base embedding}:
%In contrast to relation extraction, many methods have been proposed to connect text information and knowledge base to facilitate knowledge base embedding.
%\newcite{wang2014knowledge} combined text information and knowledge base by embedding entities and the words in their names in the same vector space.
%\newcite{neelakantan2015compositional} learns the co-occurrence based textual relation representations to help with knowledge base completion.
%\newcite{toutanova2015representing} trained continuous representations of knowledge base and textual relations jointly, which allows for deeper interactions between the sources of information and achieved significant improvement.
%The success of this joint model on knowledge base embedding inspires us to employ a similar idea on relation extraction.

\section{Background and Problem}
%!TEX root = ../naaclhlt2019.tex

The goal in the task of Relation Extraction is to predict a KB relation that holds for a pair of entities given a set of sentences mentioning them (or \emph{NA} if no such relation exists).
The input is a KB $\Psi$ with relation set $\mathcal{R}_{\Psi}$, a set of relations of interest $\mathcal{R},\ \ \mathcal{R} \subseteq \mathcal{R}_{\Psi}$, and an automatically labelled training dataset $\mathcal{D}$ obtained via distant supervision.
%(For example, in 
%(In our setting $\mathcal{R}_{\Psi}$ consists of all 23K Freebase relations while $\mathcal{R}$ is a smaller subset of relations of interest).
Given a sentence mentioning entities $h,t$, the output is a relation $r\in\mathcal{R}$ that holds for $h,t$ or the catch-all relation {\em NA} if no such $r$ exists.

\paragraph*{Knowledge Base and Distant Supervision.}

% A KB $\Psi$ is a structured representation of human-curated facts about entities of interest, often encoded as \emph{triples}.
As customary, we denote a KB $\Psi$ with relation scheme $\mathcal{R}_\Psi$ as a set of \emph{triples} $\mathcal{T}_\Psi=\{(h, r, t)\in\mathcal{E}_\Psi \times \mathcal{R}_\Psi \times \mathcal{E}_\Psi\}$, where $\mathcal{E}_{\Psi}$ is the set of entities of interest.
Distant supervision exploits the KB to automatically annotate sentences in a corpus containing mentions of entities with the relations they participate in.
Formally, a labeled dataset for relation extraction consists of fact triples $\{(h_i, r_i, t_i)\}_{i=1}^N$ and a multi-set of extracted sentences for each triple $\{\mathcal{S}_i\}_{i=1}^N$, such that
each sentence $s \in \mathcal{S}_i$ mentions both the head entity $h_i$ and the tail entity $t_i$.

\paragraph*{Problem Statement.}

Given an entity pair $(h, t)$ and a set of sentences $\mathcal{S}$ mentioning them, the RE task is to estimate the probability of each relation in $\mathcal{R} \cup \{\mathit{NA}\}$.
Formally, for each relation $r$, we want to predict $P(r \mid h,t,\mathcal{S})$.

In practice, the input set of sentences $\mathcal{S}$ can have arbitrary size. 
For the sake of computational efficiency, we normalize the set size to a fixed number $T$ by splitting large sets and oversampling small ones. 
We also restrict the length of each sentence in the set by a constant $L$ by truncating long sentences and padding short ones. 

\section{Methodology}
%!TEX root = ../naaclhlt2019.tex

We now go over the details of our framework outlined in Figure~\ref{workflow} for unifying the learning of the language and the knowledge representations used for relation extraction. 
In a nutshell, we use LSTM with attention mechanisms for language representation and we follow the approach of~\newcite{trouillon2016complex} for KB embedding.

%---------------------------------------------------------------

\subsection{Language Representation}

%We learn the language representations based on the input set of sentences for each entity pair with a neural architecture similar to RNN setting in \cite{wu2017adversarial}.

\textbf{Input Representation.}
%We represent each word in each sentence as a real-valued vector capturing lexical and semantic features pertaining to relation extraction.
%Given a word embedding matrix $W^{wrd}$ of size $d_w \times \lvert V \rvert$, we map every word $w_i$ in $s$ to a column vector $\mathbf{w}_i^d \in \mathbb{R}^{d_w}$ 
%where $V$ is the input vocabulary and $d_w$ is the size of word embedding.
%Further, we incorporate word position embeddings to capture the distances between each word to the entities mentioned in the text.
%Similarly to \cite{zeng2014relation}, each relative distance is mapped to a randomly initialized position vector in $\mathbb{R}^{d_p}$, where $d_p$ is the size of position embedding.
%For word $w_i$, we obtain the position vector $\mathbf{w}_i^p$.
%Finally, the overall embedding of word $w_i$ is $\mathbf{w}_i^E=[(\mathbf{w}_i^d)^\top, (\mathbf{w}_i^p)^\top]^\top$.
For each word token, we use pretrained word embeddings and randomly initialized position embeddings \cite{zeng2014relation} to project it into $(d_w+d_p)$-dimensional space, where $d_w$ is the size of word embedding and $d_p$ is the size of position embedding.

\noindent \textbf{Sentence Encoder.}
For each sentence $s_i$, we apply a non-linear transformation to the vector representation of $s_i$ to derive a feature vector $z_i=f(s_i;\theta)$ given a set of parameters $\theta$.
In this paper, we adopt bidirectional LSTM with $d_s$ hidden units as $f(s_i;\theta)$ \cite{zhou2016attention}.
%The network contains two sub-networks for the forward pass and the backward pass respectively.
%Here, we use element-wise sum to combine the forward and backward pass outputs.
%The output of the $i$-th word is shown in the following equation:
%
%\begin{equation}
%z_i=[\overrightarrow{z_i}\oplus \overleftarrow{z_i}]
%\end{equation}

\noindent \textbf{Multi-level Attention Mechanisms.}
We employ attention mechanisms at both word-level and sentence-level to allow the model to softly select the most informative words and sentences during training~\cite{zhou2016attention,lin2016neural}.
%
%\noindent\emph{Word-level attention}. Let $H_w=[z_1, \dots, z_L]$ be the matrix with output vectors produced at the LSTM layer; the representation of sentence $\mathbf{s}$ (truncated to length $L$) is:
%%
%\begin{gather}
%G_w =  \tanh (H_w) \\
%\alpha_w = \text{softmax}(w_w^{\top}G_w) \\
%\mathbf{s} = H\alpha_w^{\top}
%\end{gather}
%
%\noindent where $H_w \in \mathbb{R}^{d_s\times L}$, $w_w\in\mathbb{R}^{d_s\times1}$ $\alpha_w^\top\in\mathbb{R}^{L \times 1}$, $\mathbf{s}\in\mathbb{R}^{d_s\times1}$ and $w_w$ is a trained parameter vector.
%
%\noindent\emph{Sentence-level attention}. Similarly, the language representation $\mathbf{s}_L$ can be obtained from the matrix of sentence representations $H_s=[\mathbf{s}_1, \mathbf{s}_2, \dots, \mathbf{s}_T]$ as follows:
%%
%\begin{gather}
%G_s = \tanh (H_s) \\ 
%\alpha_s = \text{softmax}(w_s^\top G_s) \\
%\mathbf{s}_L = H_s \alpha_s^{\top}
%\end{gather}
%
%\noindent where $H_s \in \mathbb{R}^{d_s\times T}$, $w_w\in\mathbb{R}^{d_s\times1}$ $\alpha_w^\top\in\mathbb{R}^{T \times 1}$, $\mathbf{s}_L\in\mathbb{R}^{d_s\times1}$ and $w_s$ is a trained parameter vector.
%
% \paragraph*{Language Loss Function.}
%
%\textbf{Distribution over Language Representation.}
With the learned language representation $\mathbf{s}_L$, the conditional probability $p(r|\mathcal{S}; \Theta^{(L)})$ is computed through a \emph{softmax} layer, where $\Theta^{(L)}$ is the parameters of the model to learn language representation.
%
%\begin{equation} \label{eq:lprob}
%p(r|\mathcal{S};\Theta^{(L)}) = \text{softmax}{(W^{(L)}\mathbf{s}_L+b^{(L)})}
%\end{equation}
%\noindent where $b^{(L)}\in\mathbb{R}^{K\times1}$ is a bias vector, $W^{(L)} \in \mathbb{R}^{K\times d_s}$ is the representation matrix of relations, $K=|\mathcal{R}\cup\{\mathit{NA}\}|$ and $\Theta^{(L)}$ is the parameters of the model to learn language representation.

%---------------------------------------------------------------

\subsection{Knowledge Representation}
\label{subsection:k}

Following the score function $\phi$ and training procedure of \citet{trouillon2016complex}, we can get the knowledge representations $e_h, w_r, e_t \in \mathbb{C}^{d_k}$.
%by minimizing the negative log-likelihood of the logistic model:
%%
%\begin{equation}
%\sum_{(h, r, t) \in \Omega} \log(1+\exp(-\mathbf{Y}_{ht}^{(r)}\phi(h,r,t)))
%\end{equation}
With the knowledge representations and the scoring function, we can obtain the conditional probability $p(r|(h,t); \Theta^{(G)})$ for each relation $r$:
\begin{equation*} \label{eq:kprob}
p(r|(h, t); \Theta^{(G)})=\frac{e^{\phi(e_h, w_r, e_t)}}{\sum_{r'\in\mathcal{R}\cup\{\mathit{NA}\}}e^{\phi(e_h, w_{r'}, e_t)}}
\end{equation*}

\noindent where $\Theta^{(G)}$ corresponds to the knowledge representations $e_h, w_r, e_t \in \mathbb{C}^{d_k}$.
Since $\mathit{NA}\notin \mathcal{R}_{\Psi}$, we use a randomized complex vector as $w_{\mathit{NA}}$.

%---------------------------------------------------------------

\subsection{Connecting the Pieces}

As stated, this paper seeks an elegant way of connecting language and knowledge representations for the RE task.
In order to achieve that, we use separate loss functions (recall Figure~\ref{workflow}) to guide the language and knowledge representation learning and a third loss function that ties the predictions of these models thus nudging the parameters towards agreement.

% We found that better results were achieved if we started from stable knowledge representations. 
% In other words, we first train knowledge representations $e_h, w_r, e_t$ (as in the previous section) on the whole KB independently and then use them as the initialization point for the \emph{joint} learning of the final knowledge representations with the language representation.

%\paragraph*{Loss Functions.}

The cross-entropy losses based on the language and knowledge representations are defined as:
\begin{align} \label{eq:lloss}
\mathcal{J}_L & = -\frac1N\sum_{i=1}^N \log p(r_i|\mathcal{S}_i; \Theta^{(L)}) \\
\label{eq:kloss}
\mathcal{J}_G & = -\frac1N\sum_{i=1}^N \log p(r_i|(h_i, t_i); \Theta^{(G)})
\end{align} 

\noindent where $N$ denotes the size of the training set.
%Since the language representation is learned from local context, the subscript $L$ in $\mathcal{J}_L$ means local.
%Then the cross-entropy loss based on knowledge representations (recall Eq.~\ref{eq:kprob}) can be defined as:
%
%\begin{equation} \label{eq:kloss}
%\mathcal{J}_G = -\frac1N\sum_{i=1}^N \log p(r_i|(h_i, t_i); \Theta^{(G)})
%\end{equation} 
%\noindent Since the knowledge representation is learned from the whole KB, the subscript $G$ in $\mathcal{J}_G$ means global.
Finally, we use a cross-entropy loss to measure the dissimilarity between two distributions, thus connecting them, and formulate model learning as minimizing $\mathcal{J}_{D}$:

\begin{equation} \label{eq:closs}
%\mathcal{J}_C = \sum_{i=1}^N KL(p(\cdot|\mathcal{S}) \Vert  p(\cdot|h, t))
\mathcal{J}_D = -\frac1N\sum_{i=1}^N \log p(r_i^* |\mathcal{S}_i;\Theta^{(L)})
\end{equation}

%\noindent where $KL(p(\cdot|\mathcal{S}) \Vert  p(\cdot|h, t))$ is the KL-divergence from $p(r_i|\mathcal{S})$ to $p(r_i|h,t)$, $p(r_i|\mathcal{S})$ and $p(r_i|h,t)$ has the form of Eq. \ref{eq:lprob} and \ref{eq:kprob}.
\noindent where $r_i^*=\arg\max_{r\in\mathcal{R}\cup\{\mathit{NA}\}}p(r|(h_i, t_i);\Theta^{(G)})$.
%We also tried to use KL-divergence as $\mathcal{J}_D$ but the cross-entropy loss generally performed better.

%---------------------------------------------------------------

\subsection{Model Learning}

%\textbf{Optimization.}
Based on Eq.~\ref{eq:lloss},~\ref{eq:kloss},~\ref{eq:closs}, we form the joint optimization problem for model parameters as
\begin{equation} \label{eq:loss}
\min_{\Theta} \mathcal{J} = \mathcal{J}_L + \mathcal{J}_G + \mathcal{J}_D + \lambda\lVert \Theta \rVert_2^2
\end{equation}
\noindent where $\Theta=\Theta^{(L)} \cup \Theta^{(G)}$. % is all the parameters of the considered model.
%Collectively optimizing Eq.~\ref{eq:loss} allows heterogeneous representations enhance each other. The language representation can leverage prior knowledge existing in the whole KB but not in the training dataset $\mathcal{D}$. Also, the knowledge representations can be refined with the text information related to the facts.
% In order to solve the joint optimization problem in Eq.~\ref{eq:loss},
The knowledge representations are first trained on the whole KB independently and then used as the initialization for the \emph{joint} learning.
We adopt the stochastic gradient descent with mini-batches and Adam~\cite{kingma2014adam} to update $\Theta$, employing different learning rates $lr_1$ and $lr_2$ on $\Theta^{(L)}$ and $\Theta^{(G)}$ respectively
% Note that the language representation is learned from scratch but the knowledge representations are already pre-trained on the whole KB. 
%We employ different learning rates $lr_1$ and $lr_2$ on $\Theta^{(L)}$ and $\Theta^{(G)}$ respectively, where $lr_2$ is significantly smaller than $lr_1$.

%\paragraph*{Regularization.}

%Dropout, proposed by \newcite{hinton2012improving}, prevents co-adaptation of hidden units by randomly omitting feature detectors from the network during forward propagation. We employ both input and output dropout on LSTM layers. In addition, we constrain L2-norms for the weight vectors as shown in Eq.~\ref{eq:loss}.

%---------------------------------------------------------------

\subsection{Relation Inference}
%We now discuss the strategy for relation prediction, which is essentially the same as the one of~\newcite{weston2013connecting}. 
In order to get the conditional probability $p(r|(h, t), \mathcal{S}; \Theta)$, we use the weighed average to combine the two distribution $p(r|\mathcal{S}; \Theta^{(L)})$ and $p(r|(h, t); \Theta^{(G)})$:
\begin{equation}
\label{combine}
\begin{split}
p(r|(h,t), \mathcal{S}; \Theta) = \alpha * p(r|\mathcal{S};\Theta^{(L)}) \\ 
+ (1-\alpha) * p(r|(h,t); \Theta^{(G)}).
\end{split}
\end{equation}

\noindent where $\alpha$ is the combining weight of the weighted average. Then, the predicted relation $\hat r$ is
\begin{equation}
\hat r = \underset{r \in \mathcal{R}\cup\{\mathit{NA}\}}{\operatorname{argmax}}\ p(r|(h,t), \mathcal{S}; \Theta).
\end{equation}

\section{Experiments}
%!TEX root = ../naaclhlt2019.tex

%We now report on an experimental evaluation of our framework against the state-of-the-art for the RE task.

%\subsection{Datasets}
\paragraph*{Datasets.}

We evaluate our model on the widely used {\bf NYT} dataset~\cite{riede2010modeling} by aligning Freebase relations mentioned in the New York Times Corpus.
% The Freebase relations are divided into two parts, one for training and one for testing. 
Articles from years 2005-2006 are used for training while articles from 2007 are used for testing.
As our KB, we used a Freebase subset with the 3M entities with highest degree (i.e., participating in most relations).
%Freebase is a manually curated Web-scale KB with approximately 80M entities, 23k kinds of relations and 1.2B facts.
%While Freebase is no longer maintained, it remains an invaluable resource in this area. 
Moreover, to prevent the knowledge representation from memorizing the true relations for entity pairs in the test set, we removed all entity pairs present in the NYT.

\noindent {\bf Evaluation Protocol:}
Following previous work \cite{mintz2009distant}, we evaluate our model using held-out evaluation which approximately measures the precision without time-consuming manual evaluation. 
We report both Precision/Recall curves and Precision@N (P@N) in our experiments, ignoring the probability predicted for the {\em NA} relation.
Moreover, to evaluate each sentence in the test set as in previous methods, we append $T$ copies of each sentence into $\mathcal{S}$ for each testing sample.

\noindent {\bf Word Embeddings:}
In this paper, we used the freely available 300-dimensional pre-trained word embeddings distributed by \newcite{pennington2014glove} to help the model generalize to words not appearing in the training set.

\begin{table}[t]
\small
\begin{center}
\begin{tabular}{| l|l|l |}
\hline
learning rate on $\Theta^{(L)}$ & $lr_1$ & $5\times10^{-4}$ \\
learning rate on $\Theta^{(K)}$ & $lr_2$ & $1\times10^{-5}$ \\
size of word position embedding & $d_p$ & $25$ \\
state size for LSTM layers & $d_s$ & $320$ \\
input dropout keep probability & $p_i$ & $0.9$ \\
output dropout keep probability & $p_o$ & $0.7$ \\
L2 regularization parameter & $\lambda$ & $0.0003$\\
combining weight parameter& $\alpha$ & $0.6$ \\
\hline
\end{tabular}
\end{center}
\caption{Hyperparameter setting} \label{param}
\end{table}

\noindent{\bf Hyperparameter Settings:}
For hyperparameter tuning, we randonly sampled $10\%$ of the training set as a development set.
All the hyperparameters were obtained by evaluating the model on the development set.
With the well-tuned hyperparameter setting, we run each model five times on the whole training set and report the average P@N.
For Precision/Recall curves, we just select the results from the first run of each model.
For training, we set the iteration number over all the training data as $30$.
Values of the hyperparameters used in the experiments can be found in Table~\ref{param}.

%\begin{figure*}[t]
%\begin{center}
% \includegraphics[height=2.6in]{compare.png}
%\end{center}
%\caption{The Precision/Recall curves of variants of our proposed model.
%Left: All variants of our proposed model in the recall [0-0.3] region;
%Middle: The same plot zoomed to the recall [0-0.2] region with only {\bf \HRERE-base} and {\bf \HRERE-text};
%Right: The same plot with only {\bf \HRERE-naive} and {\bf \HRERE-full}.}
%\label{compare}
%\end{figure*}

\begin{figure}[ht]
\begin{center}
\vskip-1.25em
 \includegraphics[height=2.4in]{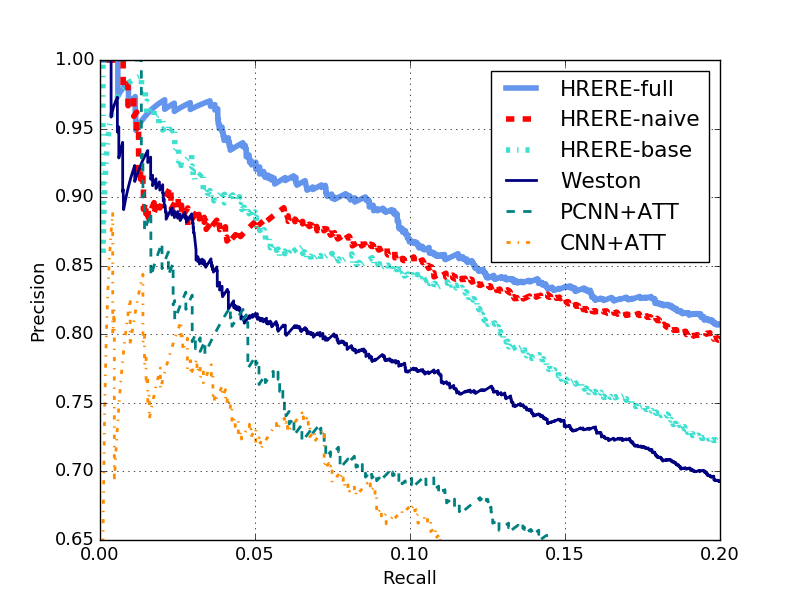}
\end{center}
\caption{The Precision/Recall curves of previous state-of-the-art methods and our proposed framework.}\label{baseline}
\end{figure}

\begin{table}[h]
\begin{center}
\begin{tabular}{l | l | l | l}
\hline
P@N(\%) & $10\%$ & $30\%$ & $50\%$ \\ \hline
Weston & $79.3$ & $68.6$ & $60.9$ \\
\HRERE-base & $81.8$ & $70.1$ & $60.7$ \\
\HRERE-naive & $83.6$& $74.4$ & $65.7$ \\
\HRERE-full & $\mathbf{86.1}$ & $\mathbf{76.6}$ & $\mathbf{68.1}$ \\

%\HRERE-base & $81.8\pm3.0$ & $70.1\pm2.2$ & $60.7\pm1.2$ \\
%\HRERE-naive & $83.6\pm2.7$& $74.4\pm3.0$ & $65.7\pm2.6$ \\
%\HRERE-full & $\mathbf{86.1\pm2.7}$ & $\mathbf{76.6\pm3.0}$ & $\mathbf{68.1\pm2.4}$ \\
\hline
\end{tabular}
\end{center}
\caption{P@N of {\bf Weston}  and variants of our proposed framework.}\label{prec}
\end{table}

\begin{table*}[ht]
\small
\begin{center}
\begin{tabular}{| l | p{8cm} | l | l | l |}
\hline
Relation & Textual Mention & base & naive & full \\ \hline
{\em contains} & Much of the {\bf middle east} tension stems from the sense that shiite power is growing, led by {\bf Iran}. & 
$0.311$ & $0.864$ & $\mathbf{0.884}$ \\ \hline
{\em place\_of\_birth} & Sometimes I rattle off the names of movie stars from {\bf Omaha}: Fred Astaire, {\bf Henry Fonda}, Nick Nolte \dots & 
$0.109$ & $0.605$ & $\mathbf{0.646}$ \\ \hline
% {\em neighborhood\_of} & Most of them also grew up in {\bf New York City} neighborhoods like Hell's Kitchen, Forest Hills, Washington Heights and {\bf Kew Gardens}, whiling away countless hours playing \dots &
% $0.196$ & $0.427$ & $\mathbf{0.946}$ \\ \hline
{\em country} & Spokesmen for {\bf Germany} and Italy in Washington said yesterday that they would reserve comment until the report is formally released at a news conference in {\bf Berlin} today. &
$0.237$ & $0.200$ & $\mathbf{0.880}$ \\ \hline
\end{tabular}
\end{center}
\caption{Some examples in NYT corpus and the predicted probabilities of the true relations.}\label{examples}
\end{table*}

%\subsection{Performance Comparison and Analysis}
\paragraph*{Methods Evaluated.}
\label{analysis}
We study three variants of our framework:
(1) {\bf \HRERE-base}: basic neural model with local loss $\mathcal{J}_L$ only;
(2) {\bf \HRERE-naive}: neural model with both local loss $\mathcal{J}_L$ and global loss $\mathcal{J}_G$ but without the dissimilarities $\mathcal{J}_D$;
(3) {\bf \HRERE-full}: neural model with both local and global loss along with their dissimilarities.
We compare against two previous state-of-the-art neural models, {\bf CNN+ATT} and {\bf PCNN+ATT} \cite{lin2016neural}.
We also implement a baseline {\bf Weston} based on the strategy following \newcite{weston2013connecting}, namely to combine the scores computed with the methods stated in this paper directly without joint learning.
%Note that both scores are computed with the methods stated in this paper instead of the original paper.

\paragraph*{Analysis.}
Figure~\ref{baseline} shows the Precision/Recall curves for all the above methods.
As one can see, {\bf \HRERE-base} significantly outperforms previous state-of-the-art neural models and {\bf Weston} over the entire range of recall.
However, {\bf \HRERE-base} performs worst compared to all other variants, while {\bf \HRERE-full} always performs best as shown in Figure~\ref{baseline} and Table~\ref{prec}.
This suggests that introducing knowledge representation consistently results in improvements, which validates our motivating hypothesis.
{\bf \HRERE-naive} simply optimizes both local and global loss at the same time without attempting to connect them.
% In a sense, this is the closest approach to the combining strategy of~\newcite{weston2013connecting}, except of course for the training procedure.
We can see that {\bf \HRERE-full} is not only consistently superior but also more stable than {\bf \HRERE-naive} when the recall is less than 0.1.
One possible reason for the instability is that the results may be dominated by one of the representations and biased toward it.
%, since there is no connection between the two heterogeneous representations. 
This suggests that (1) jointly learning the heterogeneous representations bring mutual benefits which are out of reach of previous methods that learn each independently; (2) connecting heterogeneous representations can increase the robustness of the framework.

\paragraph*{Case Study.}

% \begin{table}[ht]
% % \tiny
% \begin{center}
% % \begin{tabular}{| l | p{2.5cm} | l | l | l |}
% % \hline
% % Relation & Textual Mention & base & naive & full \\ \hline
% % {\em contains} & Much of the {\bf middle east} tension stems from the sense that shiite power is growing, led by {\bf Iran}. & 
% % $0.311$ & $0.864$ & $\mathbf{0.884}$ \\ \hline
% % {\em place\_of\_birth} & Sometimes I rattle off the names of movie stars from {\bf Omaha}: Fred Astaire, {\bf Henry Fonda}, Nick Nolte \dots & 
% % $0.109$ & $0.605$ & $\mathbf{0.646}$ \\ \hline
% % %{\em country} & Spokesmen for {\bf Germany} and Italy in Washington said yesterday that they would reserve comment until the report is formally released at a news conference in {\bf Berlin} today. & $0.237$ & $0.200$ & $\mathbf{0.880}$ \\ \hline
% % \end{tabular}
% \begin{tabular}{|ccc|}
% \hline
% \multicolumn{3}{|p{7.5cm}|}{Much of the {\bf middle east} tension stems from the sense that shiite power is growing, led by {\bf Iran}.}\\
% \hline
% \multicolumn{3}{|l|}{relation: {\em contains}} \\
% base: $0.311$ & naive: $0.864$ & \textbf{full}: $\mathbf{0.884}$\\
% \hline\hline
% \multicolumn{3}{|p{7.5cm}|}{Sometimes I rattle off the names of movie stars from {\bf Omaha}: Fred Astaire, {\bf Henry Fonda}, Nick Nolte \dots}\\
% \hline
% \multicolumn{3}{|l|}{relation: {\em place\_of\_birth}} \\
% base: $0.109$ & naive: $0.605$ & \textbf{full}: $\mathbf{0.646}$\\
% \hline
% \end{tabular}
% \end{center}
% \caption{Some examples in NYT corpus and the predicted probabilities of the true relations.}\label{examples}
% \end{table}

 Table~\ref{examples} shows two examples in the testing data.
 For each example, we show the relation, the sentence along with entity mentions and the corresponding probabilities predicted by {\bf \HRERE-base} and {\bf \HRERE-full}. 
 The entity pairs in the sentence are highlighted with bold formatting.

 From the table, we have the following observations:
 (1) The predicted probabilities of three variants of our model in the table match the observations and corroborate our analysis.
 %the analysis discussed in Section~\ref{analysis}.
 (2) From the text of the two sentences, we can easily infer that {\em middle east contains Iran} and {\em Henry Fonda was born in Omaha}.
 However, {\bf \HRERE-base} fails to detect these relations, suggesting that it is hard for models based on language representations alone to detect implicit relations, which is reasonable to expect.
 With the help of KBE, the model can effectively identify implicit relations present in the text.
(3) It may happen that the relation cannot be inferred by the text as shown in the last example.
It's a common wrong labeled case caused by distant supervision.
It is a case of an incorrectly labeled instance, a typical occurrence in distant supervision.
However, the fact is obviously true in the KBs.
As a result, {\bf \HRERE-full} gives the underlying relation according to the KBs.
This observation may point to one direction of de-noising weakly labeled textual mentions generated by distant supervision.

\section{Conclusion}
%!TEX root = ../naaclhlt2019.tex

This paper describes an elegant neural framework for jointly learning heterogeneous representations from text and from facts in an existing knowledge base.
Contrary to previous work that learn the two disparate representations independently and use simple schemes to integrate predictions from each model, we introduce a novel framework using an elegant loss function that allows the proper connection between the the heterogeneous representations to be learned seamlessly during training.
Experimental results demonstrate that the proposed framework outperforms previous strategies to combine heterogeneous representations and the state-of-the-art for the RE task.
A closer inspection of our results show that our framework enables both independent models to enhance each other.
% Our framework was tested on established benchmarks and built on publicly available datasets and obtain substantial improvements over state-of-the-art RE methods.
%We observe not only substantial improvements over state-of-the-art RE methods but also gains over the previous approach of~\newcite{weston2013connecting}.
%Furthermore, our experiments suggest an immediate line of future work, by revealing that the textual mentions are likely to be incorrectly labeled by distant supervision when the heterogeneous representations disagree with each other.
%An interesting application of our framework could be to de-noise the training corpus generated by distant supervision.
%We also believe our framework is sufficiently generic to be applicable to other NLP tasks that could be solved by integrating heterogeneous representations.

\section*{Acknowledgments}

This work was supported in part by grants from the Natural Sciences and Engineering Research Council of Canada and a gift from Diffbot Inc.

\bibliographystyle{acl_natbib}
\bibliography{ref}

\end{document}